# InverseFaceNet: Deep Monocular Inverse Face Rendering

Hyeongwoo Kim [1,2]　　Michael Zollhöfer [1,2,3]　　Ayush Tewari [1,2]
Justus Thies [4]　　Christian Richardt [5]　　Christian Theobalt [1,2]

[1] Max-Planck-Institute for Informatics　　[2] Saarland Informatics Campus
[3] Stanford University　　[4] Technical University of Munich　　[5] University of Bath

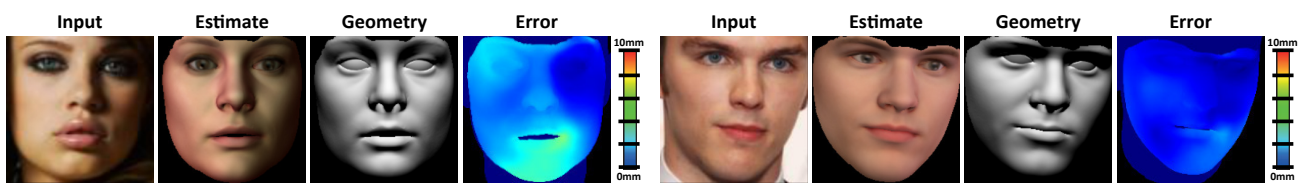

Figure 1. Our single-shot deep inverse face renderer *InverseFaceNet* obtains a high-quality geometry, reflectance and illumination estimate from just a single input image. We jointly recover the facial pose, shape, expression, reflectance and incident scene illumination. *From left to right:* the input photo, our estimated face model, its geometry, and the pointwise Euclidean geometry error compared to Garrido et al. [19].

## Abstract

*We introduce InverseFaceNet, a deep convolutional inverse rendering framework for faces that jointly estimates facial pose, shape, expression, reflectance and illumination from a single input image. By estimating all parameters from just a single image, advanced editing possibilities on a single face image, such as appearance editing and relighting, become feasible in real time. Most previous learning-based face reconstruction approaches do not jointly recover all dimensions, or are severely limited in terms of visual quality. In contrast, we propose to recover high-quality facial pose, shape, expression, reflectance and illumination using a deep neural network that is trained using a large, synthetically created training corpus. Our approach builds on a novel loss function that measures model-space similarity directly in parameter space and significantly improves reconstruction accuracy. We further propose a self-supervised bootstrapping process in the network training loop, which iteratively updates the synthetic training corpus to better reflect the distribution of real-world imagery. We demonstrate that this strategy outperforms completely synthetically trained networks. Finally, we show high-quality reconstructions and compare our approach to several state-of-the-art approaches.*

## 1. Introduction

Inverse rendering aims to reconstruct scene properties such as geometry, reflectance and illumination from image data. This reconstruction is fundamentally challenging, as it inevitably requires inverting the complex real-world image formation process. It is also an ill-posed problem as certain effects, such as low-frequency reflectance and illumination, can be indistinguishable [45]. Inverse rendering, for example, enables relighting of faces by modifying the scene illumination and keeping the face reflectance and geometry fixed.

Recently, optimization-based approaches for inverse face rendering were introduced with convincing results [2, 19, 28, 34, 60]. One of the key ingredients that enables to disentangle pose, geometry (both related to shape and facial expression), reflectance and illumination are specific priors that constrain parameters to plausible values and distributions. Formulating such priors accurately for real faces is difficult, as they are unknown a priori. The priors could be learned by applying inverse rendering to a large dataset of real face images, but this is highly challenging without having the priors a priori.

We take a different approach to solve this chicken-and-egg problem. Instead of formulating explicit priors, we directly learn inverse face rendering with a deep neural network that implicitly learns priors based on the training corpus. As annotated training data is hard to come by, we train on synthetic face images with known model parameters (geometry, reflectance and illumination). This is similar to existing approaches [46, 47, 52], but the used parameter distribution does not match that of real-world faces and environments. As a result, the learned implicit priors are rather weak and do not generalize well to in-the-wild images.

The approach of Li et al. [38] introduces a self-augmented procedure for training a CNN to regress the spatially varying surface appearance of planar exemplars. Our self-supervised bootstrapping approach extends their training strategy to handle unknown, varying geometry. In addition, we resample based on a mean-adaptive Gaussian in each bootstrapping step, which helps to populate out-of-domain samples, especially at the domain boundary.

In contrast to many other approaches, InverseFaceNet also regresses color reflectance and illumination. Our main technical contribution is the introduction of a self-supervised bootstrapping step in our training loop, which continuously updates the training corpus to better reflect the distribution of real-world face images. The key idea is to apply the latest version of the inverse face rendering network to real-world images without ground truth, to estimate the corresponding face model parameters, and then to create synthetic face renderings for perturbed, but known, parameter values. In this way, we are able to bootstrap additional synthetic training data that better reflects the real-world distribution of face model parameters, and our network therefore better generalizes to the real-world setting. Our experiments demonstrate that our approach greatly improves the quality of regressed face models for real face images compared to approaches that are trained exclusively on synthetic data.

The main contribution of our paper is InverseFaceNet – a real-time, deep, single-shot inverse face rendering network that estimates pose, shape, expression, color reflectance and illumination from just a single input image in a single forward pass, and is multiple orders of magnitude faster than previous optimization-based methods estimating similar models. To improve the accuracy of the results, we further propose a loss function that measures model-space distances directly in a modified parameter space. We further propose self-supervised bootstrapping of a synthetic training corpus based on real images without available ground truth to produce labeled training data that follows the real-world parameter distribution. This leads to significantly improved reconstruction results for in-the-wild face photos.

## 2. Related Work

**Inverse Rendering (of Faces)** The goal of inverse rendering is to invert the graphics pipeline, i.e., to recover the geometry, reflectance (albedo) and illumination from images or videos of a scene – or, in our case, a face. Early work on inverse rendering made restrictive assumptions like known scene geometry and calibrated input images [45, 65]. However, recent work has started to relax these assumptions for specific classes of objects such as faces. Deep neural networks have been shown to be able to invert simple graphics pipelines [32, 42], although these techniques are so far only applicable to low-resolution grayscale images. In contrast, our approach reconstructs full-color facial reflectance and illumination, as well as geometry. Aldrian and Smith [2] use a 3D morphable model for optimization-based inverse rendering. They sequentially solve for geometry, reflectance and illumination, while we jointly regress all dimensions at once. Thies et al. [60] recently proposed a real-time inverse rendering approach for faces that estimates a person's identity and expression using a blendshape model with reflectance texture and colored spherical harmonics illumination. Their approach is designed for reenactment and is visually convincing, but relies on non-linear least-squares optimization, which requires good initialization and a face model calibration step from multiple frames, while our approach estimates a very similar face model in a single shot, from a single in-the-wild image, in a fraction of the time. Inverse rendering has also been applied to face image editing [40, 55], for example to apply makeup [34, 35]. However, these approaches perform an image-based intrinsic decomposition without an explicit 3D face model, as in our case.

**Face Models** The appearance and geometry of faces are often modeled using 3D morphable models [5] or active appearance models [14]. These seminal face models are powerful and expressive, and remain useful for many applications even though more complex and accurate appearance models exist [30, 37]. Recently, a large-scale parametric face model [7] was created from 10,000 facial scans, Booth et al. [6] extend 3D morphable models to "in-the-wild" conditions, and deep appearance models [17] extend active appearance models by capturing geometry and appearance of faces more accurately under large unseen variations. We describe the face model we use in Section 4.

**3D Face Reconstruction** The literature on reconstructing face geometry, often with appearance, but without any illumination, is much more extensive compared to inverse rendering. We focus on single-view techniques and do not further discuss multi-view or multi-image approaches [23, 29, 44, 48, 57]. Recent techniques approach monocular face reconstruction by fitting active appearance models [1, 17], blendshape models [9, 18, 19, 61], affine face models [15, 16, 20, 46, 51, 54, 58, 62], mesh geometry [26, 33, 47, 48, 52], or volumetric geometry [24] to input images or videos. Shading-based surface refinement can extract even fine-scale geometric surface detail [11, 19, 26, 47, 48, 52]. Many techniques use facial landmark detectors for more robustness to changes in the head pose and expression, and we discuss them in the next section. A range of approaches use RGB-D input [e.g. 36, 59, 64], and while they achieve impressive face reconstruction results, they rely on depth data which is typically not available for in-the-wild images or videos.

Deep neural networks have recently shown promising results on various face reconstruction tasks. In a paper before its time, Nair et al. [42] proposed an analysis-by-synthesis algorithm that iteratively explores the parameter space of a black-box generative model, such as active appearance models (AAM) [14], to learn how to invert it, e.g., to convert a photo of a face into an AAM parameter vector. We are inspired by their approach and incorporate a self-supervised bootstrapping approach into our training process (see Section 7) to make our technique more robust to unseen inputs, in our case real photographs.

Richardson et al. [46] use iterative error feedback [12] to optimize the shape parameters of a grayscale morphable model from a single input image. Richardson et al. [47] build on this to reconstruct detailed depth maps of faces with learned shape-from-shading. Sela et al. [52] learn depth and correspondence maps directly using image-to-image translation, and follow this with non-rigid template mesh alignment. Dou et al. [16] regress only the identity and expression components of a face. All these approaches are trained entirely on synthetic data [5]. Tran et al. [62] train using a photo collection, but their focus lies on estimating morphable model parameters to achieve robust face recognition. In contrast to these approaches, ours not only recovers face geometry and texture, but a more complete inverse rendering model that also comprises color reflectance and illumination, from just a single image without the need for iteration. Jackson et al. [24] directly regress a volumetric face representation from a single input image, but this requires a large dataset with matching face images and 3D scans, and does not produce an editable face model, as in our case. Schönborn et al. [51] optimize a morphable model using Bayesian inference, which is robust and accurate, but very slow compared to our approach (taking minutes rather than milliseconds). Tewari et al. [58] learn a face regressor in a self-supervised fashion based on a CNN-based encoder and a differentiable expert-designed decoder. Our self-supervised bootstrapping approach combines the advantages of synthetic and real training data, which leads to similar quality reconstructions without the need for a hand-crafted differentiable rendering engine.

**Face Alignment** Many techniques in 3D face reconstruction, including ours, draw on facial landmark detectors for robustly identifying the location of landmark keypoints in the photograph of a face, such as the outline of the eyes, nose and lips. These landmarks can provide valuable pose-independent initialization. Chrysos et al. [13] and Jin and Tan [27] provide two recent surveys on the many landmark detection approaches that have been proposed in the literature. Perhaps unsurprisingly, deep learning approaches [4, 68] are again among the best available techniques. However, none of these techniques works perfectly [8, 56]: facial hair, glasses and poor lighting conditions pose the largest problems. In many cases, these problems can be overcome when looking at video sequences instead of single images [43], but this is a different setting to ours.

## 3. Overview

We first detect a set of 66 2D facial landmarks [50], see Figure 2. The landmarks are used to segment the face from the background, and mask out the mouth interior to effectively remove the parts of the image that cannot be explained by our model. The masked face is input to our deep inverse face rendering network (Section 6), which is trained on synthetic facial imagery (Section 5) using a parametric face and image formation model (Section 4). Starting from this low-quality corpus, we apply our self-supervised bootstrapping approach that updates the parameter distribution of the training set (Section 7) to bootstrap a training corpus that better approximates the real-world distribution. This leads to higher quality reconstructions (Section 8). Finally, we discuss limitations (Section 8.4) and conclude (Section 9).

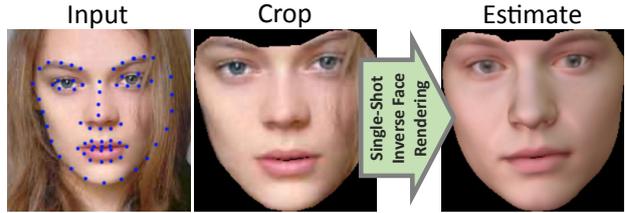

Figure 2. Our single-shot inverse face renderer regresses a dense reconstruction of the pose, shape, expression, skin reflectance and incident illumination from a single photograph.

## 4. The Space of Facial Imagery

We parameterize face images using $m = 350$ parameters:

$$\boldsymbol{\theta} = \left(\mathbf{R}, \boldsymbol{\theta}^{[\text{s}]}, \boldsymbol{\theta}^{[\text{e}]}, \boldsymbol{\theta}^{[\text{r}]}, \boldsymbol{\theta}^{[\text{i}]}\right) \in \mathbb{R}^m. \quad (1)$$

Here, $\mathbf{R}$ specifies the global rotation (3 parameters), $\boldsymbol{\theta}^{[\text{s}]}$ the shape (128), $\boldsymbol{\theta}^{[\text{e}]}$ the expression (64), $\boldsymbol{\theta}^{[\text{r}]}$ the skin reflectance (128), and $\boldsymbol{\theta}^{[\text{i}]}$ the incident illumination (27). Note that we do not include translation as our network works on consistently segmented input images (see Figure 2 and Section 3).

### 4.1. Affine Face Model

We employ an affine face model to parameterize facial geometry $\mathcal{F}^{[\text{g}]} \in \mathbb{R}^{3V}$ and reflectance $\mathcal{F}^{[\text{r}]} \in \mathbb{R}^{3V}$, where $V$ is the number of vertices of the underlying manifold template mesh. The geometry vector $\mathcal{F}^{[\text{g}]}$ stacks the $V$ 3D coordinates that define the mesh's embedding in space. Similarly, the reflectance vector $\mathcal{F}^{[\text{r}]}$ stacks the RGB per-vertex reflectance values. The space of facial geometry is modeled by the shape $\boldsymbol{\theta}^{[\text{s}]} \in \mathbb{R}^{N_\text{s}}$ and expression $\boldsymbol{\theta}^{[\text{e}]} \in \mathbb{R}^{N_\text{e}}$ parameters:

$$\mathcal{F}^{[\text{g}]}(\boldsymbol{\theta}^{[\text{s}]}, \boldsymbol{\theta}^{[\text{e}]}) = \mathbf{a}^{[\text{g}]} + \sum_{i=1}^{N_\text{s}} \mathbf{b}_i^{[\text{s}]} \sigma_i^{[\text{s}]} \theta_i^{[\text{s}]} + \sum_{j=1}^{N_\text{e}} \mathbf{b}_j^{[\text{e}]} \sigma_j^{[\text{e}]} \theta_j^{[\text{e}]}. \quad (2)$$

The spatial embedding is modeled by a linear combination of orthonormal basis vectors $\mathbf{b}_i^{[\text{s}]}$ and $\mathbf{b}_j^{[\text{e}]}$, which span the shape and expression space, respectively. $\mathbf{a}^{[\text{g}]} \in \mathbb{R}^{3V}$ is the average geometry of a neutral expression, the $\sigma_i^{[\text{s}]}$ are the shape standard deviations and the $\sigma_j^{[\text{e}]}$ are the standard deviations of the expression dimensions.

Per-vertex reflectance is modeled similarly using a small number of reflectance parameters $\boldsymbol{\theta}^{[\text{r}]} \in \mathbb{R}^{N_\text{r}}$:

$$\mathcal{F}^{[\text{r}]}(\boldsymbol{\theta}^{[\text{r}]}) = \mathbf{a}^{[\text{r}]} + \sum_{i=1}^{N_\text{r}} \mathbf{b}_i^{[\text{r}]} \sigma_i^{[\text{r}]} \theta_i^{[\text{r}]}. \quad (3)$$

Here, $\mathbf{b}_i^{[\text{r}]}$ are the reflectance basis vectors, $\mathbf{a}^{[\text{r}]}$ is the average reflectance and the $\sigma_i^{[\text{r}]}$ are the standard deviations.

The face model is computed from 200 high-quality 3D scans [5] of Caucasians (100 male and 100 female) using

PCA. We use the $N_s = N_r = 128$ most significant principal directions to span our face space. The used expression basis is a combination of the Digital Emily model [3] and FaceWarehouse [10] (see Thies et al. [60] for details). We use PCA to compress the over-complete blendshapes (76 vectors) to a subspace of $N_e = 64$ dimensions.

## 4.2. Image Formation

We assume the face to be *Lambertian*, illumination to be distant and smoothly varying, and there is no self-shadowing. We thus represent the incident illumination on the face using second-order spherical harmonics (SH) [41, 45]. Therefore, the irradiance at a surface point with normal $\mathbf{n}$ is given by

$$\mathcal{B}(\mathbf{n} \,|\, \boldsymbol{\theta}^{[\text{i}]}) = \sum_{k=1}^{b^2} \boldsymbol{\theta}_k^{[\text{i}]} H_k(\mathbf{n}), \quad (4)$$

where $H_k$ are the $b^2 = 3^2 = 9$ SH basis functions, and the $\boldsymbol{\theta}_k^{[\text{i}]}$ are the corresponding illumination coefficients. Since we consider colored illumination, the parameters $\boldsymbol{\theta}_k^{[\text{i}]} \in \mathbb{R}^3$ specify RGB colors, leading to $3 \cdot 9 = 27$ parameters in total.

We render facial images based on the SH illumination using a full perspective camera model $\Pi : \mathbb{R}^3 \to \mathbb{R}^2$. We render the face using a mask (painted once in a preprocessing step) that ensures that the rendered facial region matches the crops produced by the 66 detected landmark locations (see Figure 2). The global rotation of the face is modeled with three Euler angles using $\mathbf{R} = \text{Rot}_{xyz}(\alpha, \beta, \gamma)$ that successively rotate around the $x$-axis (up, $\alpha$), $y$-axis (right, $\beta$), and $z$-axis (front, $\gamma$) of the camera-space coordinate system.

## 5. Initial Synthetic Training Corpus

Training our deep inverse face rendering network requires ground-truth training data $\{\mathbf{I}_i, \boldsymbol{\theta}_i\}_{i=1}^N$ in the form of corresponding pairs of image $\mathbf{I}_i$ and model parameters $\boldsymbol{\theta}_i$. However, training on real images is challenging, since the ground-truth parameters cannot easily be obtained for a large dataset. We therefore train our network based on synthetically rendered data, where exact ground-truth labels are available.

We sample $N = 200{,}000$ parameter vectors $\boldsymbol{\theta}_i$ and use the model described in Section 4 to generate the corresponding images $\mathbf{I}_i$. Data generation can be interpreted as sampling from a probability $P(\boldsymbol{\theta})$ that models the distribution of real-world imagery. However, sampling from this distribution is in general difficult and non-trivial. We therefore assume statistical independence between the components of $\boldsymbol{\theta}$, i.e.,

$$P(\boldsymbol{\theta}) = P(\mathbf{R}) P(\boldsymbol{\theta}^{[\text{s}]}) P(\boldsymbol{\theta}^{[\text{e}]}) P(\boldsymbol{\theta}^{[\text{r}]}) P(\boldsymbol{\theta}^{[\text{i}]}). \quad (5)$$

This enables us to efficiently generate a parameter vector $\boldsymbol{\theta}$ by independently sampling each subset of parameters.

We uniformly sample the yaw and pitch rotation angles $\alpha, \beta \sim \mathcal{U}(-40°, 40°)$ and the roll angle $\gamma \sim \mathcal{U}(-15°, 15°)$ to reflect common head rotations. We sample shape and reflectance parameters from the Gaussian distributions provided by the parametric PCA face model [5]. Since we already scale with the appropriate standard deviations during face generation (see Equations 2 and 3), we sample both from a standard normal distribution, i.e., $\boldsymbol{\theta}^{[\text{s}]}, \boldsymbol{\theta}^{[\text{r}]} \sim \mathcal{N}(0, 1)$. The expression basis is based on artist-created blendshapes that only approximate the real-world distribution of the space of human expressions; this will be addressed by the self-supervised bootstrapping presented in Section 7. We thus uniformly sample the expression parameters using $\boldsymbol{\theta}^{[\text{e}]} \sim \mathcal{U}(-12, 12)$. To prevent closing the mouth beyond anatomical limits, we apply a bias of 4.8 to the distribution of the first parameter[1]. Finally, we sample the illumination parameters using $\boldsymbol{\theta}^{[\text{i}]} \sim \mathcal{U}(-0.2, 0.2)$, except for the constant coefficient $\boldsymbol{\theta}_1^{[\text{i}]} \sim \mathcal{U}(0.6, 1.2)$ to account for the average image brightness, and set all RGB components to the same value. The self-supervised bootstrapping step presented in Section 7 automatically introduces colored illumination.

## 6. InverseFaceNet

Given the training data $\{\mathbf{I}_i, \boldsymbol{\theta}_i\}_{i=1}^N$ consisting of $N$ images $\mathbf{I}_i$ and the corresponding ground-truth parameters $\boldsymbol{\theta}_i$, we train a deep inverse face rendering network $\mathcal{F}$ to invert image formation. In the following, we provide details on our network architecture and the employed loss function.

### 6.1. Network Architecture

We have tested several different networks based on the popular AlexNet [31] and ResNet [21] architectures, both pre-trained on ImageNet [49]. In both cases, we resize the last fully-connected layer to match the dimensionality of our model (350 outputs), and initialize biases with 0, and weights $\sim \mathcal{N}(0, 0.01)$. These minimally modified networks provide the baseline we build on. We propose more substantial changes to the training procedure by introducing a novel model-space loss in Section 6.2, which more effectively trains the same network architecture. The color channels of the input images are normalized to the range $[-0.5, 0.5]$ before feeding the data to the network. We show a comparison between the results of AlexNet and ResNet-101 in Section 8.1, and thus choose AlexNet for our results.

**Input Pre-Processing** The input to our network is a color image of a masked face with a resolution of $240 \times 240$ pixels (see Figure 2). We mask the face to remove any background and the mouth interior, which cannot be explained by our face model. For this, we use detected landmarks [50] and resize their bounding box uniformly to fit inside $240 \times 240$ pixels, to approximately achieve scale and translation invariance.

**Training** We train all our inverse face rendering networks using the Caffe deep learning framework [25] with stochastic gradient descent based on AdaDelta [66]. We perform 75K batch iterations with a batch size of 32 for training our baseline approaches. To prevent overfitting, we use an $\ell_2$-regularizer (*aka* weight decay) of 0.001. We train with a base learning rate of 0.01.

---

[1]The first parameter mainly corresponds to mouth opening and closing.

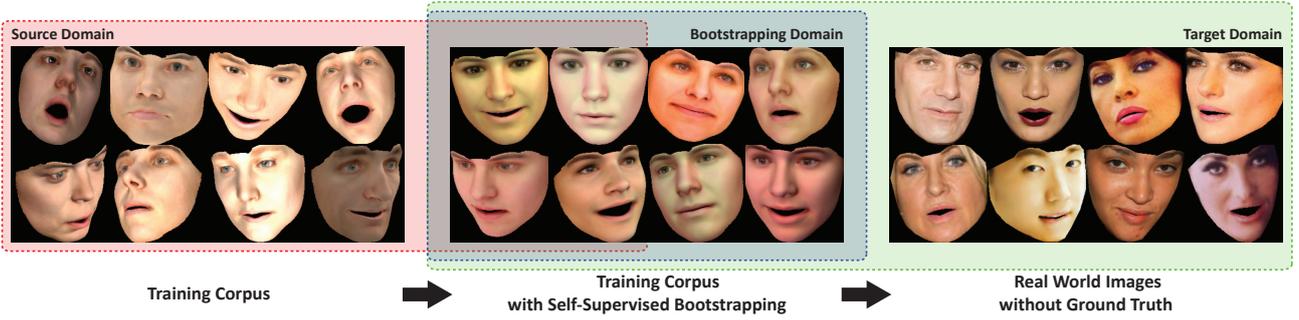

Figure 3. Our approach updates the initial training corpus (left) based on real-world images without available ground truth (right) using a self-supervised bootstrapping approach. The generated new training corpus (middle) better matches the real-world face distribution.

## 6.2. Model-Space Parameter Loss

We use a weighted norm to define a model-space loss between the predicted parameters $\boldsymbol{\theta}$ and ground-truth $\boldsymbol{\theta}_g$ by taking the statistics of the face model into account:

$$\mathcal{L}(\boldsymbol{\theta}, \boldsymbol{\theta}_g) = \|\boldsymbol{\theta} - \boldsymbol{\theta}_g\|_{\mathbf{A}}^2 \quad (6)$$

$$= (\boldsymbol{\theta} - \boldsymbol{\theta}_g)^\top \underbrace{\mathbf{A}}_{\boldsymbol{\Sigma}^\top \boldsymbol{\Sigma}} (\boldsymbol{\theta} - \boldsymbol{\theta}_g). \quad (7)$$

Here, $\boldsymbol{\Sigma}$ is a weight matrix that incorporates the standard deviations $\boldsymbol{\sigma}^{\bullet}$ of the different parameter dimensions:

$$\boldsymbol{\Sigma} = \mathrm{diag}(\omega_{\mathbf{R}} \mathbf{1}_3, \omega_{\mathrm{s}} \boldsymbol{\sigma}^{[\mathrm{s}]}, \omega_{\mathrm{e}} \boldsymbol{\sigma}^{[\mathrm{e}]}, \omega_{\mathrm{r}} \boldsymbol{\sigma}^{[\mathrm{r}]}, \omega_{\mathrm{i}} \mathbf{1}_{27}) \in \mathbb{R}^{m \times m}. \quad (8)$$

The coefficients $\omega_{\bullet}$ balance the global importance of the different groups of parameters, and $\mathbf{1}_k$ is a $k$-dimensional vector of ones. We use the same values $(\omega_{\mathbf{R}}, \omega_{\mathrm{s}}, \omega_{\mathrm{e}}, \omega_{\mathrm{r}}, \omega_{\mathrm{i}}) = (400, 50, 50, 100, 20)$ for all our results. Note that we do not scale the rotation and illumination dimensions individually. Intuitively speaking, our model-space loss enforces that the first PCA coefficients (higher variation basis vectors) should match the ground truth more accurately than the later coefficients (lower-variation basis vectors), since the former have a larger contribution to the final 3D geometry and skin reflectance of the reconstructed face in model space (see Equations 2 and 3). As shown in Section 8, this leads to more accurate reconstruction results. The difference to Zhu et al. [68] is the computation of the weights, which leads to a statistically meaningful metric.

## 7. Self-Supervised Bootstrapping

The real-world distribution of the model parameters $\boldsymbol{\theta}$ is in general unknown for in-the-wild images $\mathbf{I}_{\mathrm{real}}$. Until now, we have sampled from a manually prescribed probability distribution $P(\boldsymbol{\theta})$, which does not exactly represent the real-world distribution. The goal of the self-supervised bootstrapping step is to make the training data distribution better match the real-world distribution of a corpus $\mathcal{R}$ of in-the-wild face photographs. To this end, we automatically bootstrap the parameters for the training corpus. Note that this step is unsupervised and does not require the ground-truth parameters for images in $\mathcal{R}$ to be available.

**Algorithm 1** Self-Supervised Bootstrapping
1: $\mathcal{F} \leftarrow$ train_network_on_synthetic_faces();
2: $\mathcal{R} \leftarrow$ corpus_of_real_images();
3: **for** (number of bootstrapping steps $N_{\mathrm{boot}}$) **do**
4: $\quad \boldsymbol{\theta}_r \leftarrow$ inverse_rendering($\mathcal{R}, \mathcal{F}$); ▷ (step 1)
5: $\quad \boldsymbol{\theta}_r' \leftarrow$ resample_parameters($\boldsymbol{\theta}_r$); ▷ (step 2)
6: $\quad \mathcal{R}' \leftarrow \{$generate_images($\boldsymbol{\theta}_r'$), $\boldsymbol{\theta}_r'\}$; ▷ (step 3)
7: $\quad \mathcal{F} \leftarrow$ continue_training($\mathcal{F}, \mathcal{R}'$); ▷ (step 4)
8: **end for**

### 7.1. Bootstrapping

Bootstrapping based on uniform resampling with replacement $\mathbf{I}_r \sim P(\mathbf{I}) = 1/N$ cannot solve the problem of mismatched distributions. Hence, we propose a domain-adaptive approach that resamples new proposals from a mean-adaptive Gaussian distribution based on real images:

$$P(\mathbf{I}_r(\boldsymbol{\theta}) \mid \mathbf{I}_{\mathrm{real}}) \sim \boldsymbol{\theta}(\mathbf{I}_{\mathrm{real}}) + \mathcal{N}(\mathbf{0}, \boldsymbol{\sigma}^2), \quad (9)$$

where $\mathbf{I}_r(\boldsymbol{\theta})$ is the deterministic rendering process, we compute the inverse of the rendering process $\boldsymbol{\theta}(\mathbf{I}_{\mathrm{real}})$ using InverseFaceNet, and $\mathcal{N}(\cdot)$ is a noise distribution. This shifts the distribution closer to the target distribution of real images $\mathbf{I}_{\mathrm{real}}$. Moreover, adding a non-zero variance $\boldsymbol{\sigma}^2 > \mathbf{0}$ populates out-of-domain samples especially at the domain boundary. Our approach takes the network of the last bootstrapping iteration as final output, instead of averaging the intermediate networks. This prevents from being biased to the manually prescribed sampling distribution of earlier training stages.

### 7.2. Algorithm

Our self-supervised parameter bootstrapping is a four-step process (see Algorithm 1). It starts with a deep neural network $\mathcal{F}$ initially trained on a synthetic training corpus (see Section 5) for 15K batch iterations. This guarantees a suitable initialization for all weights in the network. Given a set of images from the corpus of real-world images $\mathcal{R}$, we first obtain an estimate of the corresponding model parameters $\boldsymbol{\theta}_r$, i.e., $\boldsymbol{\theta}(\mathbf{I}_{\mathrm{real}})$ in Equation 9, using the synthetically trained network (step 1). These reconstructed parameters are used to seed the bootstrapping. In step 2, we apply small perturbations to the reconstructed parameters based on the noise distribution $\mathcal{N}(\mathbf{0}, \boldsymbol{\sigma}^2)$. This generates new data around the seed points in model space, and allows the network to slowly adapt to the real-world parameter distribution. We use the following to re-sample the pose, shape, expression, reflectance and illumination parameters, generating two perturbed parameter vectors for each reconstruction: $\alpha,\beta,\gamma: \mathcal{U}(-5°,5°)$, $\boldsymbol{\theta}^{[s]}: \mathcal{N}(0,0.05)$, $\boldsymbol{\theta}^{[r]}: \mathcal{N}(0,0.2)$, $\boldsymbol{\theta}^{[e]}: \mathcal{N}(0,0.1)$, and $\boldsymbol{\theta}^{[i]}: \mathcal{N}(0,0.02)$. In step 3, we generate new synthetic training images $\mathbf{I}_r$ based on the resampled parameters $\boldsymbol{\theta}'_r$, i.e., $\boldsymbol{\theta}(\mathbf{I}_{\text{real}}) + \mathcal{N}(\mathbf{0},\boldsymbol{\sigma}^2)$. The result is a new synthetic training set $\mathcal{R}'$ that better reflects the real-world distribution of model parameters. Finally, the network $\mathcal{F}$ is fine-tuned for $N_{\text{iter}} = 7.5K$ batch iterations on the new training corpus (step 4). In total, we repeat this process for $N_{\text{boot}} = 8$ self-supervised bootstrapping steps.

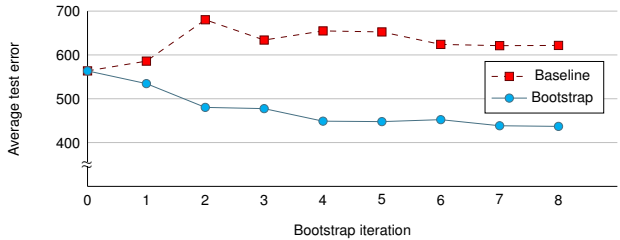

Figure 4. Model-space parameter loss (Equation 7) for the baseline and bootstrapping approaches on a synthetic test corpus with higher parameter variation than the used training corpus. While our domain-adaptive bootstrapping approach, based on a high-variation training corpus without available ground truth, continuously decreases in loss, the baseline network fails to generalize.

Over the iterations, the data distribution of the training corpus adapts and better reflects the real-world distribution of the provided in-the-wild facial imagery, as illustrated in Figure 3. We also evaluate the parameter loss throughout bootstrapping iterations in Figure 4, and observe a clear reduction with our self-supervised bootstrapping. This leads to higher quality results at test time, as shown in Section 8. The variance $\boldsymbol{\sigma}^2$ could be adaptively scaled based on the photometric error of estimates. However, we found empirically that our framework works well with a fixed variance.

## 8. Experiments and Results

We evaluate our InverseFaceNet on several publicly available datasets. We validate our design choices regarding network architecture, model-space loss, and self-supervised bootstrapping. We then show quantitative and qualitative results and comparisons on the datasets *LFW* (Labeled Faces in the Wild) [22], *300-VW* (300 Videos in the Wild) [53], *CelebA* [39], *FaceWarehouse* [10], *Volker* [63] and *Thomas* [18]. For more results, we refer to our supplemental document and video[2].

**Error Measures** We compute the **photometric error** using the RMSE of RGB pixel values (within the mask of the input image) between the input image and a rendering of the reconstructed face model. An error of 0 is a perfect color match, and 255 is the difference between black and white (i.e.

[2]Project page: http://gvv.mpi-inf.mpg.de/projects/InverseFaceNet

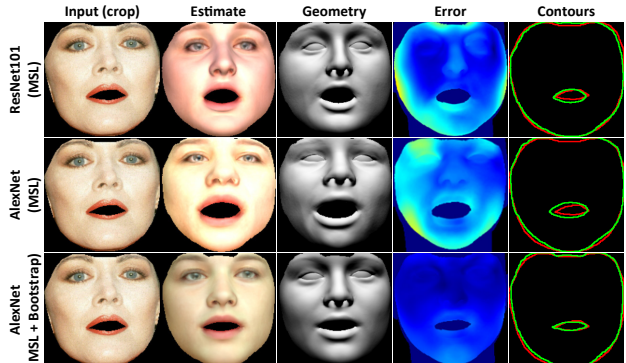

Figure 5. Qualitative comparison of ResNet-101 [21] and AlexNet [31] applied to inverse face rendering, both with model-space loss (MSL): ResNet-101 produces lower geometric error (see heatmap) while AlexNet has lower photometric error (also on average, see Table 1). AlexNet with MSL and bootstrapping clearly improves the reconstruction of reflectance and geometry, in all error categories.

lower is better). The **geometric error** measures the RMSE in mm between corresponding vertices in our reconstruction and the ground-truth geometry. We quantify the image-space overlap of the estimated face model and the input face image using the **intersection over union** (IOU) of face masks (e.g. see 'contours' in Figure 5). An IOU of 0% means no overlap, and 100% means perfect overlap (i.e. higher is better).

### 8.1. Evaluation of Design Choices

Table 1 evaluates different design choices on a test dataset of 5,914 images (one shown in Figure 5) from *CelebA* [39] using the error measures described earlier (using our implementation of Garrido et al. [19] as ground-truth geometry, up to blendshape level).

**Network Architecture** We first compare the results of the AlexNet [31] and ResNet-101 [21] architectures, both with our model-space loss (see Section 6). Reconstructions using ResNet-101 have smaller geometric errors, but worse photometric error and IOU than AlexNet, which is exemplified by Figure 5. ResNet-101 is significantly deeper than AlexNet, so training takes about $10\times$ longer and testing about $5\times$ longer. We thus use AlexNet for our inverse face rendering network, which only requires 3.9 ms for the forward pass (on an Nvidia Titan Xp). Landmark detection takes 4.5 ms and face morphing 1 ms (on the GPU). In total, our approach requires 9.4 ms.

**Importance of Model-Space Loss** Table 1 shows that our model-space loss improves on baseline AlexNet [31] in all error categories, particularly the photometric error and IOU. As our model-space loss does not modify the network architecture, the time for the forward pass remains the same fast 3.9 ms as before.

**Importance of Self-supervised Bootstrapping** Our self-supervised bootstrapping (see Section 7) significantly improves the reconstruction quality and produces the lowest errors in all categories, as shown in Table 1. This can also be seen in Figure 5, which shows plausible reconstruction of appearance and geometry, the lowest geometric errors, and improved contour overlap for our network with bootstrapping. Note that the training time for self-supervised bootstrapping includes all steps (see Algorithm 1), in particular reconstructing 100K face models (0.25 h), rendering 200K synthetic faces (2.8 h) and training for 7.5K iterations (0.5 h) for each of the 8 bootstrapping iterations (on an Nvidia GeForce GTX Titan). AlexNet with bootstrapping significantly outperforms ResNet-101 without bootstrapping in reconstruction quality, training time and test time. Note that our approach is better than Tewari et al. [58] in terms of geometry and overlap, and worse in terms of the photometric error on this test set.

Table 1. Quantitative architecture comparison, model-space parameter loss and our bootstrapping step on 5,914 test images from *CelebA* [39]. The best values for each column are highlighted in bold. Training time includes all steps except the initial training data generation. Test times are averaged over 5K images. Training on a GTX Titan and testing on a Titan Xp. Errors show means and standard deviations. *For bootstrapping, we first train 15K iterations on normal synthetic face images (see Section 5), and then bootstrap for 60K iterations (see Section 7). InverseFaceNet (AlexNet [31] with model-space loss and bootstrapping) produces the best geometric error and intersection over union.

| Approach | Training iterations | Training time [h] | Test time [ms / image] | Photometric error [8 bits] | Geometric error [mm] | Intersection over union [%] |
|---|---|---|---|---|---|---|
| AlexNet [31] | 75K | 4.14 | **3.9** | 46.26 ± 12.42 | 2.91 ± 0.99 | 90.44 ± 3.81 |
| + model-space loss | 75K | 4.36 | **3.9** | 39.71 ± 9.86 | 2.77 ± 1.00 | 92.51 ± 2.59 |
| + bootstrap (= **InverseFaceNet**) | 75K* | 29.40 | **3.9** | 34.03 ± 7.56 | **2.11** ± **0.84** | **93.96** ± **2.08** |
| ResNet-101 [21] + model-space loss | 150K | 40.99 | 21.0 | 41.23 ± 10.58 | 2.54 ± 0.87 | 92.07 ± 2.87 |
| MoFA [58] | — | — | **3.9** | **17.23** ± **4.42** | 3.94 ± 1.34 | 84.20 ± 4.23 |

Table 2. Quantitative evaluation of the geometric accuracy on 180 meshes of the FaceWarehouse [10] dataset.

| | Our approach | | Other approaches | | |
|---|---|---|---|---|---|
| | **Bootstrap** | Baseline | Garrido et al. [19] | Tewari et al. [58] | MonoFit (*see text*) |
| Error | 2.11 mm | 2.33 mm | **1.59** mm | 2.19 mm | 2.71 mm |
| SD | 0.46 mm | 0.47 mm | **0.30** mm | 0.54 mm | 0.52 mm |

### 8.2. Quantitative Evaluation

We compare the geometric accuracy of our approach to state-of-the-art monocular reconstruction techniques in Figure 6. As ground truth, we use the high-quality stereo reconstructions of Valgaerts et al. [63]. Compared to Thies et al. [60], our approach obtains similar quality results, but without the need for explicit optimization. Therefore, our approach is two orders of magnitude faster (9.4 ms vs 600 ms) than optimization-based approaches. Note that while Thies et al. [60] run in real time for face tracking, it requires significantly longer to estimate all model parameters from an initialization based on the average model. In contrast to the state-of-the-art learning-based methods by Richardson et al. [46, 47], Jackson et al. [24] and Tran et al. [62], ours obtains a reconstruction of all dimensions, including pose, shape, expression, and colored skin reflectance and illumination.

In addition, we performed a large quantitative ground-truth comparison on the FaceWarehouse [10] dataset, see Table 2. We show the mean error (in mm) and standard deviation (SD) for 180 meshes (9 different identities, each with 20 different expressions). As can be seen, our bootstrapping approach increases accuracy. Our approach is only slightly worse than the optimization-based approach of Garrido et al. [19], while being orders of magnitude faster. Bootstrapping is on par with the weakly supervised approach of Tewari et al. [58], which is trained on real images and landmarks. We also compare to a baseline network 'MonoFit' that has been directly trained on the monocular fits of Garrido et al. [19] on the *CelebA* [39] dataset. Our self-supervised bootstrapping approach obtains higher accuracy results.

### 8.3. Qualitative Evaluation

We next compare our reconstruction results qualitatively to current state-of-the-art approaches. Figure 7 compares our reconstruction to optimization-based approaches that fit a parametric face model [19] or a person-specific template mesh [18]. Our learning-based approach is significantly faster (9.4 ms vs about 2 minutes [19]), and orthogonal to optimization-based approaches, since it can be used to provide a good initial solution.

In Figure 8, we also compare to the state-of-the-art deep-learning-based approaches by Richardson et al. [46, 47], Sela et al. [52], Jackson et al. [24], Tran et al. [62] and Tewari et al. [58]. We obtain high-quality results in 9.4 ms. Most of the other approaches are slower, do *not* estimate colored skin reflectance and illumination [24, 46, 47, 52], do *not* regress the facial expressions [62], or suffer from geometric shrinking artifacts [58]. Note, we compare to Richardson et al.'s 'CoarseNet' [47], which corresponds to their earlier method [46], and estimates pose, shape and expression, followed by a model-based optimization of monochrome reflectance and illumination. We also compare to Sela et al.'s aligned template mesh. We don't compare to 'FineNet' [47] or 'fine detail reconstruction' [52] as these estimate a refined depth map/mesh, and we are interested in comparing the reconstructed parametric face models.

Figure 9 shows several monocular reconstruction results obtained with our InverseFaceNet. As can be seen, our approach obtains good estimates of all model parameters.

### 8.4. Limitations

We propose a solution to the highly challenging problem of inverse face rendering from a single image. Similar to previous learning-based approaches, ours has a few limitations. Our approach does not perfectly generalize to inputs that are outside of the training corpus. Profile views of the head are problematic and hard to reconstruct, even if they are part of the training corpus. Note that even state-of-the-art landmark trackers often fail in this scenario. Handling these cases robustly remains an open research question. Incorrect landmark localization might produce inconsistent input to our network, which harms the quality of the regressed face model. This could be addressed by more sophisticated face detection algorithms, or by joint learning of landmarks and reconstruction. Occlusions of the face, such as hair, beards, sun glasses or hands, can also be problematic. To handle these situations robustly, our approach could be trained in an occlusion-aware manner by augmenting our training corpus with artificial occlusions, similar to Zhao et al. [67].

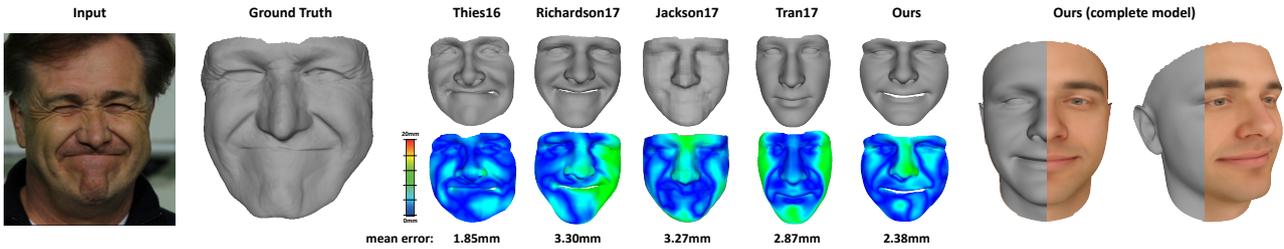

Figure 6. Quantitative comparison of geometric accuracy compared to Thies et al. [60], Richardson et al. [47], Jackson et al. [24] and Tran et al. [62] on *Volker* [63]. The heat maps visualize the pointwise Hausdorff distance (in mm) between the input and the ground-truth. The ground-truth has been obtained by the high-quality binocular reconstruction approach of Valgaerts et al. [63].

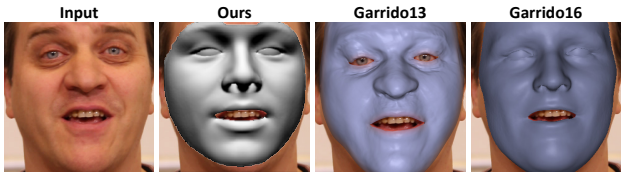

Figure 7. Qualitative comparison to optimization-based approaches [18, 19] on *Thomas* [18]. For more, see our supplemental document.

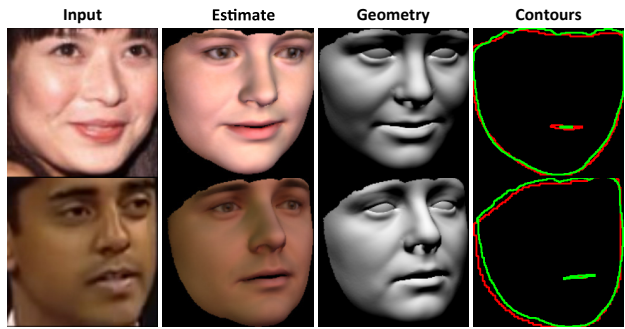

Figure 9. Qualitative results on *LFW* [22] and *300-VW* [53]. Top to bottom: input image, our estimated face model and geometry, and contours (red: input mask, green: ours). Our approach achieves high-quality reconstructions from just a single input image. For more results, we refer to the supplemental document.

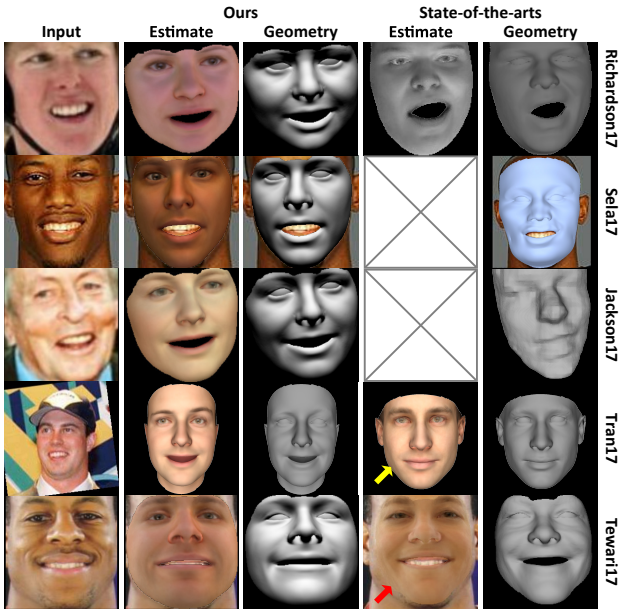

Figure 8. Comparison to a wide range of state-of-the-art learning-based approaches. From top to bottom: Comparison to Richardson et al. [47], Sela et al. [52], Jackson et al. [24], Tran et al. [62] and Tewari et al. [58]. We obtain high-quality results in 9.4 ms. Most other approaches are significantly slower, do not estimate colored skin reflectance and illumination (empty box), do not regress facial expressions (yellow arrow), or suffer from geometric shrinking (red arrow). Images from *LFW* [22], *300-VW* [53], *CelebA* [39] and Face-Warehouse [10]. For more results, see our supplemental document.

## 9. Conclusion

We have presented InverseFaceNet – a single-shot inverse face rendering framework. Our key contribution is to overcome the lack of well-annotated image datasets by self-supervised bootstrapping of a synthetic training corpus that captures the real-world distribution. This enables high-quality face reconstruction from just a single monocular image. Our evaluation shows that our approach compares favorably to the state of the art. InverseFaceNet could be used to quickly and robustly initialize optimization-based reconstruction approaches close to the global minimum. We hope that our approach will stimulate future work in this exciting field.

**Acknowledgments.** We thank True-VisionSolutions Pty Ltd for kindly providing the 2D face tracker, and we thank Aaron Jackson, Anh Tran, Mata Sela and Elad Richardson for the comparisons. This work was supported by ERC Starting Grant CapReal (335545), RCUK grant CAMERA (EP/M023281/1) and the Max Planck Center for Visual Computing and Communications (MPC-VCC).

# InverseFaceNet: Deep Monocular Inverse Face Rendering
## — Supplemental Material —

Hyeongwoo Kim [1, 2]   Michael Zollhöfer [1, 2, 3]   Ayush Tewari [1, 2]
Justus Thies [4]   Christian Richardt [5]   Christian Theobalt [1, 2]

[1] Max-Planck-Institute for Informatics   [2] Saarland Informatics Campus
[3] Stanford University   [4] Technical University of Munich   [5] University of Bath

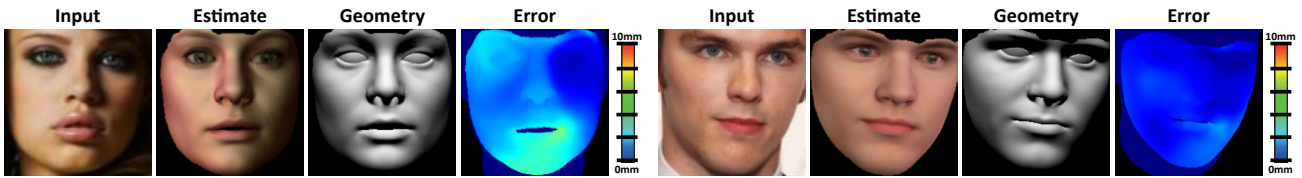

Figure 1. Our single-shot deep inverse face renderer *InverseFaceNet* obtains a high-quality geometry, reflectance and illumination estimate from just a single input image. We jointly recover the facial pose, shape, expression, reflectance and incident scene illumination. *From left to right:* the input photo, our estimated face model, its geometry, and the pointwise Euclidean geometry error compared to Garrido et al. [3].

This document provides additional discussion, comparisons and results for our approach. We discuss technical details on how we evaluate the self-supervised bootstrapping approach using a synthetic test set in Section 1. Also, we provide quantitative and qualitative comparisons in Sections 2 and 3, respectively, to further demonstrate the accuracy and effectiveness of our approach. Finally, we demonstrate the robustness of our approach on a wide range of challenging face images in Section 4.

## 1. Self-Supervised Bootstrapping

To evaluate the strength of our self-supervised bootstrapping step in the training loop, we use synthetic validation images, as it is difficult to acquire the ground-truth parameters for real-world images. This section explains in more detail the evaluation shown in Section 7.2 and Figure 4 of the main document, in particular the image sets used for training, bootstrapping and validation.

We first generate a set of 50,000 training images with a parameter distribution that has little variation; the mouth, for instance, is not opening much. We then modify the distribution with a bias and more variation in face expression and color to simulate real-world images, and generate two sets of 5,000 images each for bootstrapping and validation. The difference between the image sets is clearly visible in Figure 2.

In this evaluation, InverseFaceNet uses a set of 5,000 images without the corresponding parameters for self-supervised bootstrapping. The initialization, used weights and number of training iterations are explained in the main document. For evaluation, we visualize the face parameters estimated from premature to fully domain-adapted networks, i.e., along the bootstrapping iterations, in the testing phase as shown in Figure 3. In addition, we compute the model-space parameter loss of the validation image set. With the visual and numeric metrics, the performance of bootstrapped InverseFaceNet is compared against a vanilla AlexNet without bootstrapping. The decrease of the model-space loss via bootstrapping substantiates that the parameter distribution of the training set is automatically adapted to better match the image set used for bootstrapping, i.e., more mouth opening is added to the initial training set. This is in contrast to the regressed face with a closed mouth, and non-decreasing model-space parameter error by the baseline method, which estimates the best possible parameters only within the initial training set. On the basis of this evaluation, we conclude that our self-supervised bootstrapping approach results in better generalization to unseen input images in the real-world scenario. For an evaluation on real-world face images, we refer to the main document.

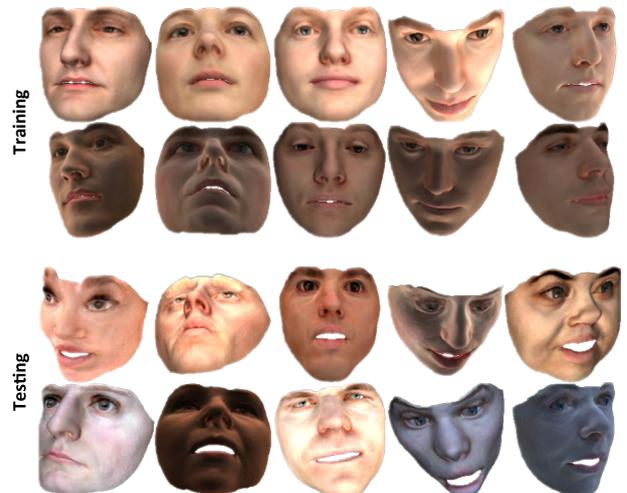

Figure 2. Images used for training (top) and testing (bottom) in the bootstrapping evaluation. The synthetic examples for testing and bootstrapping are sampled from a wider distribution than the training images. Thus, there is more variation in face shape, expression and color, such as mouth opening and colored illumination.

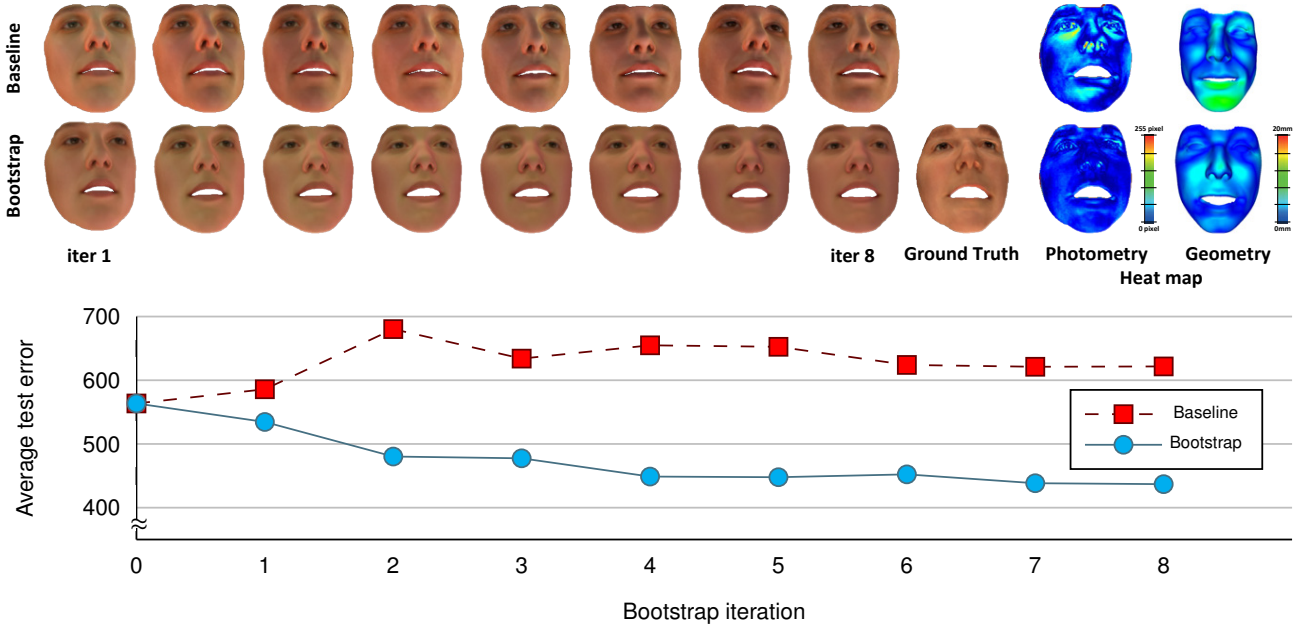

Figure 3. Comparison of baseline and bootstrapping approaches on a synthetic test corpus with higher parameter variation than in the used training corpus (also synthetic). **Top:** Reconstructions of an unseen input image after different numbers of bootstrapping iterations. Notice how the reconstructions with bootstrapping gradually converge towards the ground-truth face model (right), e.g., opening the mouth, while the baseline approach does not improve visibly over time. The last two columns visualize the photometric (2× scaled) and geometric errors at the final bootstrapping step. The mean photometric error is 16.74 pixels in the $L_1$-norm distance for the baseline method, and 12.11 pixels after bootstrapping. The Hausdorff distance is 2.56 mm and 2.01 mm for the baseline method and after bootstrapping, respectively. **Bottom:** Model-space parameter loss for the baseline and bootstrapping approaches. While our domain-adaptive bootstrapping approach continuously decreases the error by adapting the parameter distribution based on a higher variation corpus without available ground truth, the baseline network overfits to the training data and fails to generalize to the unseen data.

## 2. Additional Quantitative Evaluation

In addition to the quantitative evaluation on *FaceWarehouse* [1] in the main document, we here evaluate and compare our approach on a challenging video sequence (300 frames of *Volker* [13]). As ground-truth geometry, we use the high-quality binocular reconstructions of Valgaerts et al. [13]. Our approach outperforms Tewari et al. [11] on this sequence, and comes close to the optimization-based results of Garrido et al. [3], which is orders of magnitude slower than our approach (2 minutes vs our 9.4 ms).

Table 1. Quantitative evaluation of the geometric accuracy on 300 frames of the *Volker* dataset [13].

|       | **Ours** | Garrido et al. [3] | Tewari et al. [11] |
|-------|----------|--------------------|--------------------|
| Error | 2.10 mm  | 1.96 mm            | 2.94 mm            |
| SD    | 0.42 mm  | 0.35 mm            | 0.28 mm            |

## 3. Qualitative Evaluation

In the following, we show additional results and comparisons that unfortunately did not fit into the limited space of the main document. Specifically, we compare to the approaches of Richardson et al. [8], Sela et al. [9], Jackson et al. [5], Tran et al. [12], Tewari et al. [11], Garrido et al. [2] and Garrido et al. [3] on a variety of challenging face datasets, including *LFW* (Labeled Faces in the Wild) [4], *300-VW* (300 Videos in the Wild) [10], *CelebA* [6], *FaceWarehouse* [1], *Volker* [13] and *Thomas* [2]. The results are shown in Figure 4 to Figure 9.

We compare to the results of Richardson et al.'s 'CoarseNet' [8] and Sela et al.'s aligned template mesh [9]

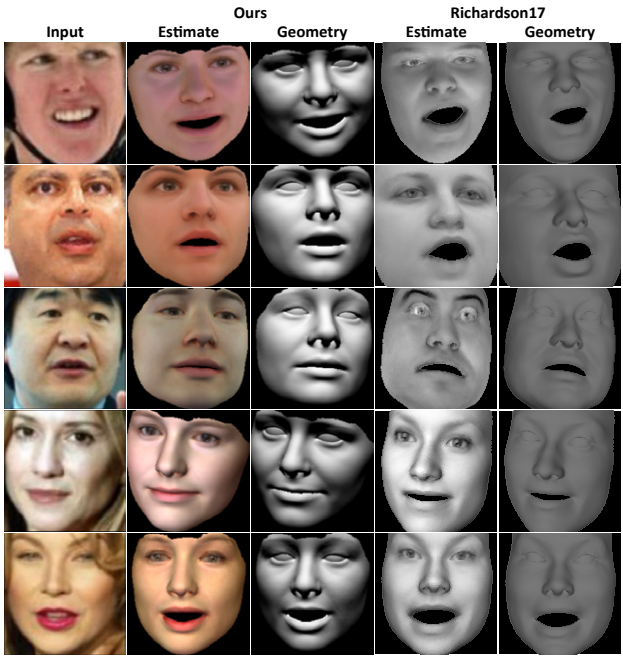

Figure 4. Qualitative comparison to Richardson et al. [8] on *LFW* [4]. Note that our reconstruction results are colored, and better fit the face shape and mouth expressions of the input images.

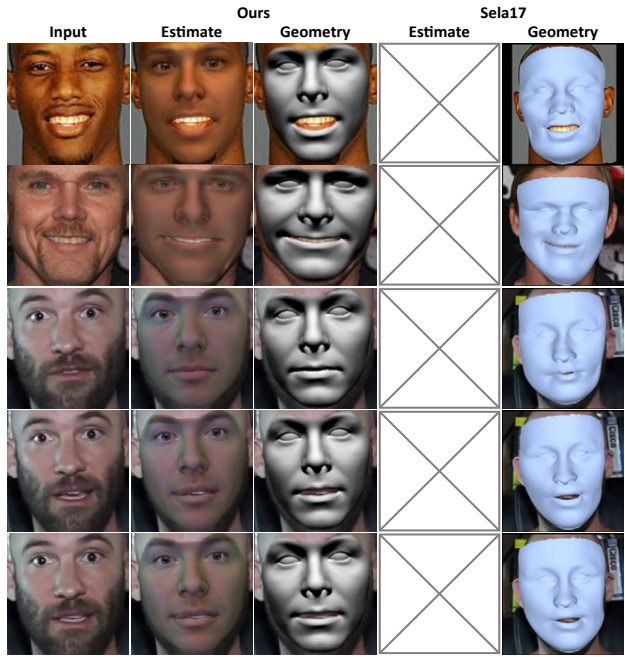

Figure 5. Qualitative comparison to Sela et al. [9] on *CelebA* [6] (top 2 rows) and *300-VW* [10] datasets. From top to bottom: our approach reconstructs facial reflectance and reliable shape, while theirs does not recover reflectance or illumination, and suffers from global shape distortion.

as we are interested in comparing the reconstructed parametric face models. As can be seen in Figures 4 and 5, we obtain similar or even higher quality results than these two state-of-the-art approaches. Note that their approaches do not require landmarks for initial cropping, but they are significantly slower due to their iterative regression strategy [8] or the involved non-rigid registration [9], and do not recover color reflectance. In contrast, our approach provides a one-shot estimate of all face model parameters.

Jackson et al. [5] recover coarse volumetric reconstructions, and do not reconstruct facial appearance or illumination (Figure 6). In contrast to Richardson et al. [7, 8] and Jackson et al. [5], our approach obtains an estimate of the colored skin reflectance and illumination.

The approach of Tran et al. [12] is targeted at face recognition, and thus does not recover the facial expression and illumination (Figure 7). Our results are comparable to Tewari et al. [11], but we avoid the geometric shrinking seen in Figure 8. Notice that their estimated geometry is visibly thinner than the input faces. Our approach also obtains similar quality results (Figure 9) as the optimization-based approaches by Garrido et al. [2, 3], while being several orders of magnitude faster. For a detailed discussion, we refer the reader to the main document.

## 4. Additional Results

Our approach works well even for the challenging images shown in Figure 10 with different head orientations (rows one and two), challenging expressions (rows three to five), and variation in skin reflectance (rows four to six). Our approach provides perceptually more plausible reconstructions due to our novel model-space loss and the self-supervised bootstrapping that automatically adapts the parameter distribution to match the real world. For more results and a detailed discussion, we refer the reader to the main document.

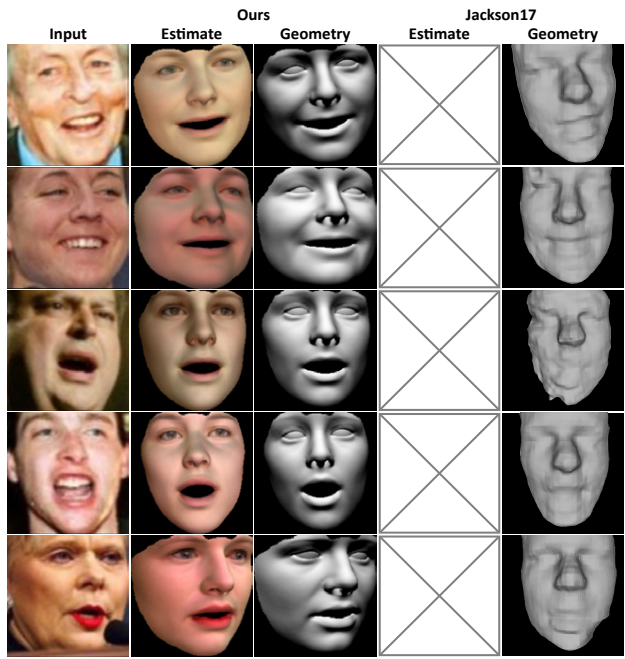

Figure 6. Qualitative comparison to Jackson et al. [5] on *LFW* [4]. Our reconstruction results include reflectance and illumination, and better fit the face shape.

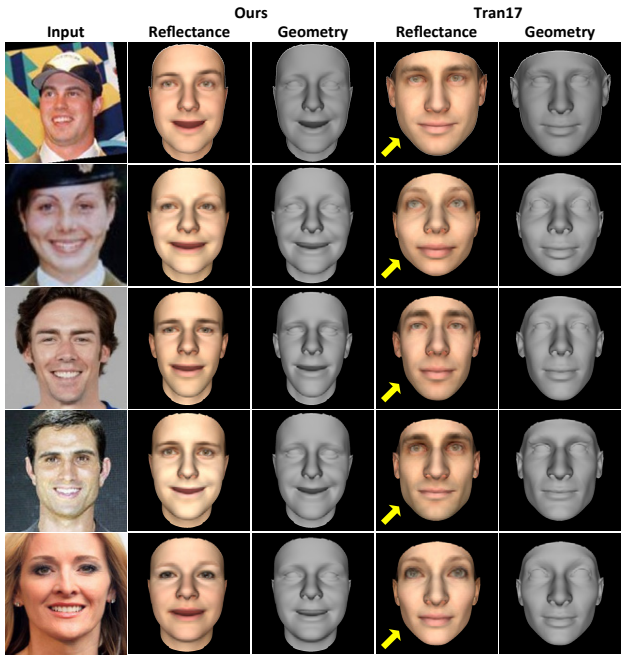
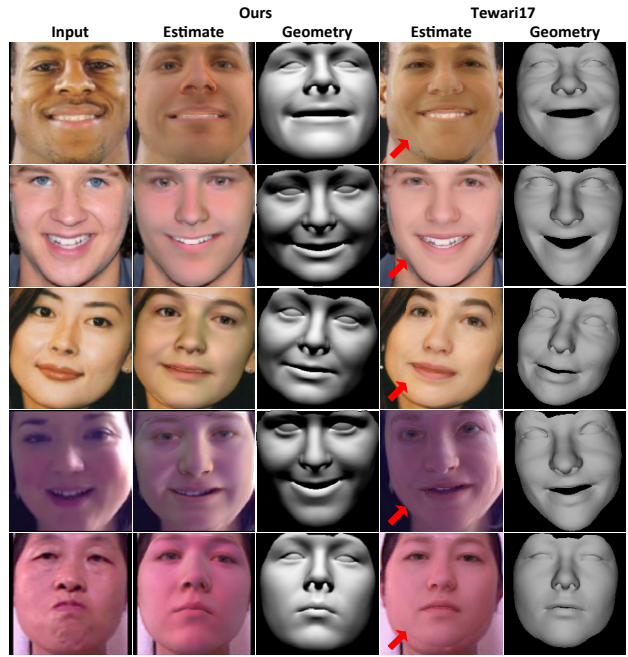

Figure 7. Qualitative comparison to Tran et al. [12] on images of the *CelebA* [6] (top 3 rows) and *LFW* [4] (rest) datasets: our approach reconstructs expressions, while theirs cannot recover this dimension (arrows).

Figure 8. Qualitative comparison to Tewari et al. [11]: our approach reconstructs the facial outline accurately, while theirs suffers from shrinking artifacts (arrows).

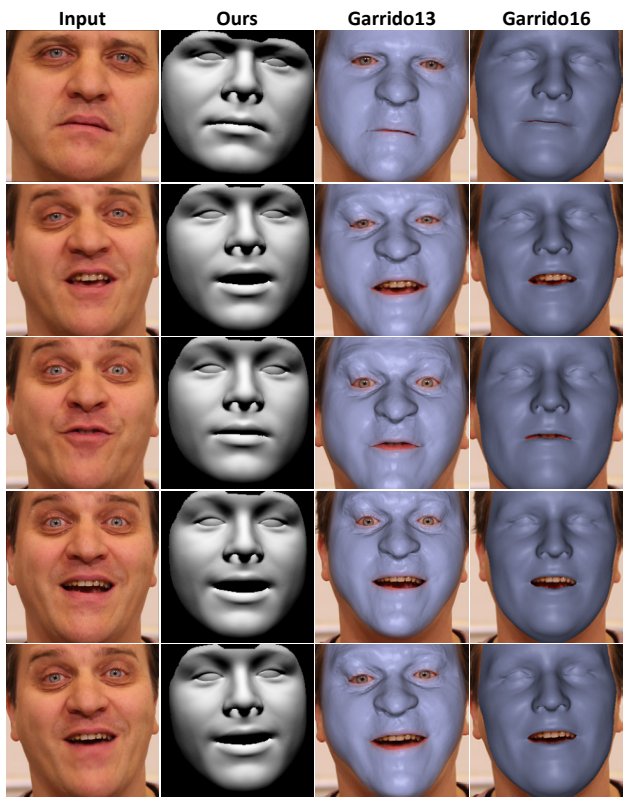

Figure 9. Qualitative comparison to optimization-based approaches [2, 3] on the *Thomas* dataset [2].

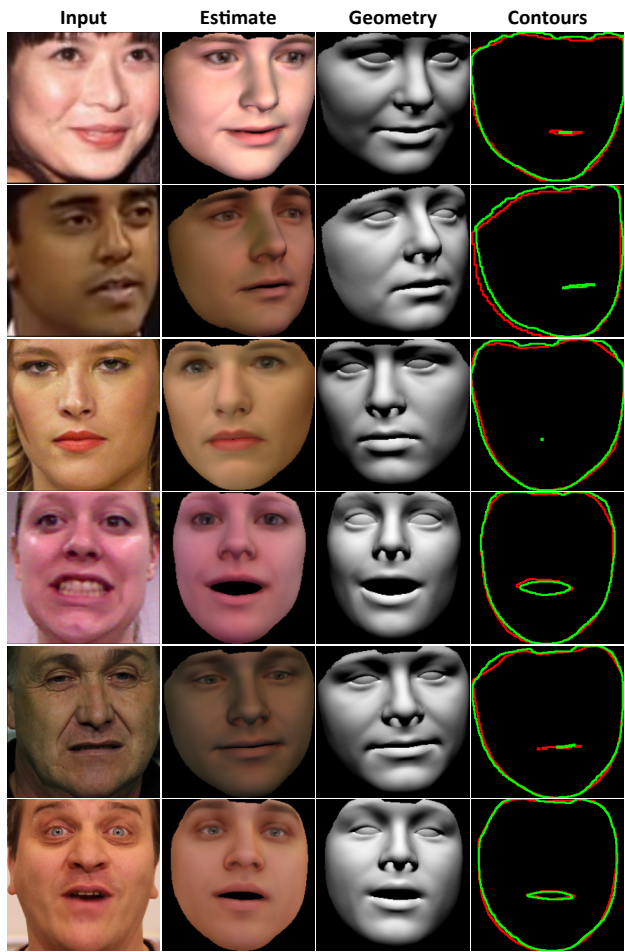

Figure 10. Qualitative results on several datasets. Left to right: input image, our estimated face model and geometry, and contours (red: input mask, green: ours). Top to bottom: *LFW* [4], *300-VW* [10], *CelebA* [6], *FaceWarehouse* [1], *Volker* [13] and *Thomas* [2]. Our approach achieves high-quality reconstructions of geometry as well as skin reflectance from just a single input image.

5